\documentclass[submit]{bioRxiv}

\usepackage{tabularx,multirow}

\newcolumntype{P}[1]{>{\raggedright\arraybackslash}p{#1}}
\newcommand{\vertrowspan}[2]{\vbox{\hbox{\multirow{#1}{*}{\rotatebox[origin=c]{90}{#2}}}}}

\usepackage{amsmath,amsfonts,bm}

\newcommand{\newterm}[1]{{\bf #1}}

\def\eqref#1{equation~\ref{#1}}

\def\1{\bm{1}}

\def\rvx{{\mathbf{x}}}
\def\rvy{{\mathbf{y}}}
\def\rvz{{\mathbf{z}}}

\def\vx{{\bm{x}}}
\def\vy{{\bm{y}}}
\def\vz{{\bm{z}}}

\def\mX{{\bm{X}}}

\def\mZ{{\bm{Z}}}

\DeclareMathAlphabet{\mathsfit}{\encodingdefault}{\sfdefault}{m}{sl}
\SetMathAlphabet{\mathsfit}{bold}{\encodingdefault}{\sfdefault}{bx}{n}

\newcommand{\parents}{Pa} 

\newcommand{\dooperator}{{\rm do}}
\newcommand{\noise}{\boldsymbol{\epsilon}} 
\newcommand{\f}{f}            
\newcommand{\indep}{\perp\!\!\!\!\!\perp}

\begin{document}

\leadauthor{Lange}

\title{Causality in the human niche: lessons for machine learning}
\shorttitle{Human-niche causality in ML}

\author[1,2,3]{Richard D. Lange \orcidlink{0000-0002-1429-8333}}
\author[4,5]{Konrad P. Kording \orcidlink {0000-0001-8408-4499}}
\affil[1]{Dept. of Computer Science, Rochester Institute of Technology}
\affil[2]{Cognitive Science PhD Program, Rochester Institute of Technology}
\affil[3]{Center for Vision Science, University of Rochester}
\affil[4]{Dept. of Neurobiology, University of Pennsylvania}
\affil[5]{CIFAR LMB}
\date{}

\maketitle

\begin{abstract}
Humans interpret the world around them in terms of cause and effect and communicate their understanding of the world to each other in causal terms. These causal aspects of human cognition are thought to underlie humans' ability to generalize and learn efficiently in new domains, an area where current machine learning systems are weak. Building human-like causal competency into machine learning systems may facilitate the construction of effective and interpretable AI. Indeed, the machine learning community has been importing ideas on causality formalized by the Structural Causal Model (SCM) framework, which provides a rigorous formal language for many aspects of causality and has led to significant advances. However, the SCM framework fails to capture some salient aspects of human causal cognition and has likewise not yet led to advances in machine learning in certain critical areas where humans excel. We contend that the problem of causality in the ``human niche'' --- for a social, autonomous, and goal-driven agent sensing and acting in the world in which humans live --- is quite different from the kind of causality captured by SCMs. For example, everyday objects come in similar types that have similar causal properties, and so humans readily generalize knowledge of one type of object (cups) to another related type (bowls) by drawing causal analogies between objects with similar properties, but such analogies are at best awkward to express in SCMs. We explore how such causal capabilities are adaptive in, and motivated by, the human niche. By better appreciating properties of human causal cognition and, crucially, how those properties are adaptive in the niche in which humans live, we hope that future work at the intersection of machine learning and causality will leverage more human-like inductive biases to create more capable, controllable, and interpretable systems.
\end{abstract}


\begin{corrauthor}
koerding@gmail.com
\end{corrauthor}

\setcounter{tocdepth}{2}
\tableofcontents

\section{Introduction}

Humans instinctively reason about the world in terms of causal relations, construct causal narratives to explain events, and communicate those relations to one another \citep{gopnikWordsThoughtsTheories1997,woodwardMakingThingsHappen2005,woodwardInterventionistTheoriesCausation2007,danksPsychologyCausalPerception2009,waldmannCausalReasoningIntroduction2017}. 
This affinity for causal thinking goes so far that humans will wrongly attribute causality simply to obtain workable causal models \citep{sumnerAssociationExaggerationHealth2014}. Such a strong bias towards causal interpretation may be a mistake in human cognition. Alternatively, it may be that humans' proclivity for causality is highly adaptive for the kinds of problem humans solve on a daily basis. More and more, machine learning (ML) and artificial intelligence (AI) systems are deployed to solve the same kinds of problems. By studying aspects of human environments, goals, and constraints --- the human niche --- we can better appreciate how causal cognition in humans is adapted to the human niche, and we can gain insights into how to build similarly efficient intelligent agents.

When expressed in terms of causally relevant variables, the world is remarkably simple. Most of the things that happen in our lives have one or a small number of definable causes. Here is a complicated but physically accurate story: When a vase falls and shatters, its kinetic energy is rapidly converted into localized stress upon impact, exceeding the material's fracture toughness and causing brittle failure through crack propagation in a network of countless interacting molecules, breaking what was once a rigid object into hundreds of smaller objects with diverse shapes and sizes. Here is a simpler and more human story: The vase at home broke when the cat knocked it off the shelf. Many such physical phenomena have one clear cause when expressed in terms of agents and their actions, and their effects can be understood in practical terms. Learning, inference, and planning are all massively facilitated in the kind of sparse causal world we live in. All of this raises the question of how we can build artificial agents that take advantage of these properties.

To reason formally about causality in AI and ML, most of the field conceptually follows the structual causal model (SCM) framework  \citep{pearlCausality2009,petersElementsCausalInference2017,kaddourCausalMachineLearning2022}. This framework conceptualizes the world as a directed acyclic graph consisting of nodes that denote variables and edges that denote causal influences. SCMs provide a precise, formal, and unifying language for a variety of specific problems within the umbrella of causality, including causal discovery, causal identification, confounding, interventions, and counterfactuals. The SCM framework also provides tools for connecting causal models to various types of data, such as Pearl's \emph{do-calculus}, which forms a bridge between observational data and interventional models \citep{pearlCausality2009}. Like any framework, modeling the world using SCMs involves making certain assumptions about the structure of the world and the data-generating process, and it provides powerful tools for learning and reasoning causally, \emph{as long as the world conforms to those assumptions}. However, many of the common assumptions within the SCM framework, while well suited for certain kinds of problem, are unnecessary or even counterproductive when considering the kinds of causal competencies required of an agent in the human niche.

Given the growing list of success stories in ML driven by domain-general models and large-scale, why should ML researchers care about causality in general and look in particular to human cognition for solutions? First and foremost, it has been proven that when it comes to causality, scale is not enough \citep{pearlCausality2009,petersElementsCausalInference2017,xiaCausalNeuralConnectionExpressiveness2021}. The reason is \newterm{confounding}, or the possibility that unobserved factors drive observed correlations; Does smoking cause lung cancer or is there a latent trait that predisposes people to both the desire to smoke and the risk of cancer? Students who spend the least time on problem sets tend to receive the highest grades, but does this mean that struggling students should spend less time, or that these metrics are confounded by the student's aptitude for the subject? Famously, answering such questions (yes, the smoking-cancer association is causal / no, the time-grade association is not) requires either domain knowledge from an expert to adjudicate which causal models are plausible, or it requires \emph{qualitatively} different kinds of data, such as learning from interventions or quasi-experiments \citep{liuQuantifyingCausalityData2021}, not just a greater quantity of data. The second reason ML researchers should draw inspiration from human cognition is that many current ML systems are still plagued by a lack of robustness to domain shift, especially when target domains have limited data. In stark contrast to this, humans excel at transferring knowledge across domains, even forming and deploying ``zero-shot'' causal models in novel situations, a powerful capacity known as causal induction. The world supplies an endless supply of novelty, and artificial systems cannot hope to gather direct experience from the long tail of possible events or contexts. The third reason ML researchers should draw inspiration from human cognition is that massive domain-general models are notoriously difficult to interpret or explain. Applying principles from human cognition to designing ML systems has the potential to make ML systems more interpretable.

The remainder of this paper is structured as follows: Section \ref{sec:human-niche} is our main story, working backwards from properties of the human niche to the aspects of human causal cognition that are adaptive to those niche properties. Along the way, we refer to work in both cognitive science and existing work in AI/ML. Section \ref{sec:scms} then revisits current practice in ML and gives a critical review of the SCM framework. We then offer some concluding remarks in Section \ref{sec:discussion}. While our analysis of the vast literature on causality in cognitive science and AI/ML is not exhaustive, we hope that it sparks useful discussions and cross-pollination of ideas across fields.

\section{Causality in the human niche}\label{sec:human-niche}

Human cognition is adapted to the environment in which humans evolved and the types of problem they need to solve --- collectively, the ``human niche.'' The concept of an ecological niche extends beyond the environment in which an organism lives. It also includes the role of an organism within its environment and the reciprocal influences that the organism has on its environment \citep{pochevilleEcologicalNicheHistory2015}. In essence, a niche encompasses an organism's physical and social environment, goals, and constraints. Our discussion in the following subsections motivates aspects of causality in human cognition by relating them to these properties of the human niche. In ML, carefully analyzing the ``niche'' where a model is deployed is crucial when designing good behaviors and inductive biases for that model. It follows that models deployed in contexts analogous to the human niche will benefit from analogous inductive biases and cognitive characteristics.

The focus of this paper is on how the human niche specifically shapes \emph{causal} aspects of human cognition. But first, what is causality? This question must necessarily be answered prior to a mathematical formalism for causality. We adopt the perspective that causality is fundamentally about agency, or the ability of agents such as humans, animals, or robots to make autonomous decisions to intervene on entities in the world and effect change. The ability to intervene is central to Woodward's definition of causality \citep{woodwardMakingThingsHappen2005,woodwardInterventionistTheoriesCausation2007}. A causal model is then any model in which actions and interventions by an agent are distinct from observations. For instance, in the SCM formalism, $p(\vy|\vx)$ is distinct from $p(\vy|\dooperator(\vx))$ \citep{pearlCausality2009} --- the do operator is what separates correlation from causality. However, we emphasize that any formalism which distinguishes agency and intervention from passive observation is a kind of causal framework, including Causal Bayes Nets \citep{pearlCausality2009}, Causal Programs \citep{tavaresLanguageCounterfactualGenerative2021}, Potential Outcomes \cite{imbensCausalInferenceStatistics2015}, and others.

We further take causality to be operational, contextual, and subjective. An agent in a deterministic but complex and chaotic world can benefit from modeling the world \emph{as if} actions are autonomous and their effects are probabilistic, and the model itself may depend on the goals, perspective, or actions of the agent. One can decide (or not) to flip a light switch, and the effect is (usually) to toggle the lights on or off. One can inadvertently knock a glass to the floor and the effect is (usually) that it shatters. From the agent's perspective, all of these examples are usefully \emph{modeled as} probabilistic outcomes conditioned on an intervention even if the outcomes are predetermined by their initial conditions, even if influences are symmetric, and regardless of concerns about free-will or whether actions themselves are independent of the environment. Fundamentally, our stance is that a formalism for causality should take the perspective of a finite and bounded agent, rather than the perspective of an omniscient oracle. This makes modeling causality a cognitive problem and one of subjective belief of the agent --- whether and how it is useful for an agent to treat the world as if it has certain causal properties --- not a matter of modeling the world as it truly is \textit{per se}.

To organize our discussion, we categorize properties of the human niche into three broad headings: properties of the environment (Section \ref{sec:niche-environment}), constraints that limit what humans and other agents can do (Section \ref{sec:niche-constraints}), and properties of the kinds of goals that humans pursue (Section \ref{sec:niche-goals}). Table \ref{tbl:niche-cognition} summarizes the properties of the human niche (middle column) and associated aspects of human causal cognition that can be seen as adaptive to those properties (right column).

\begin{table}[ht]
    \centering
    \begin{tabular}{c|c|P{6.5cm}|P{7.5cm}}
      & \textbf{Sec.} & \textbf{Property of the human niche} & \textbf{Properties of human causal cognition}\\
      \hline
      \vertrowspan{16}{Environment} 
            & \ref{sec:niche-environment-agency} & Humans have agency & Modeling is in service of doing; learning by intervening \\ \cline{2-4}
            & \ref{sec:niche-environment-complex} & The environment is complex & ``Good-enough'' models; knowledge, schemas, concepts \\ \cline{2-4}
            & \ref{sec:niche-environment-open-ended} & The environment is open-ended & Exploration and curiosity; instantiate new causal models on the fly \\ \cline{2-4}
            & \ref{sec:niche-environment-confounded} & The environment is confounded & Epistemic humility; hypothesis-driven exploration \\ \cline{2-4}
            & \ref{sec:niche-environment-designed} & The environment is designed by humans & The environment is designed for humans; human-like inductive biases are doubly useful \\ \cline{2-4}
            & \ref{sec:niche-environment-others} & Other agents exist & learning by demonstration; model others' intent \\ \cline{2-4}
            & \ref{sec:niche-environment-hierarchical} & The environment has hierarchical structure & Coarse-graining in space and time; mechanistic explanations \\ \cline{2-4}
            & \ref{sec:niche-environment-ontology} & The environment has ontological structure & Causal induction from properties (causal theories) \\ \cline{2-4}
            & \ref{sec:niche-environment-sparse} & Interactions are sparse & Focus on sparse (dyadic) interactions \\ \cline{2-4}
            & \ref{sec:niche-environment-stream} & Stream of data & Instantiate new causal models on the fly; episodic memory \\ \cline{2-4}
            & \ref{sec:niche-environment-arrow-of-time} & Arrow of time & Temporal order as a cue to causal order \\ \hline
      \vertrowspan{9}{Constraints}
            & \ref{sec:niche-constraints-observations} & Observations are partial & Represent uncertainty over both variables and models; hypothesis-driven exploration \\ \cline{2-4}
            & \ref{sec:niche-constraints-infeasible-intervention} & Interventions are sometimes costly & Mental simulation; causal induction \\ \cline{2-4}
            & \ref{sec:niche-constraints-resources} & Limited attention and working memory & Focus on sparse (dyadic) interactions; coarse-graining in space and time \\ \cline{2-4}
            & \ref{sec:niche-constraints-reasoning-slow} & Reasoning is slow & Inference over both variables and models is amortized \\ \cline{2-4}
            & \ref{sec:niche-constraints-models-wrong} & All models are wrong & Represent uncertainty over models; learning by counterfactual reasoning \\ \hline
      \vertrowspan{8}{Goals}
            & \ref{sec:niche-goals-compositional} & Goals are compositional and coarse-grained & Coarse-graining in space and time; compositional reward \\ \cline{2-4}
            & \ref{sec:niche-goals-contextual} & Goals are contextual and change over time & Focus on relevant properties/schemas; instantiate new causal models on the fly \\ \cline{2-4}
            & \ref{sec:niche-goals-intrinsic-value} & Intrinsic value of understanding & Curiosity and hypothesis-driven exploration \\ \cline{2-4}
            & \ref{sec:niche-goals-social-cooperation} & Social dynamics: cooperation, competition & model others' intent; learning by demonstration \\ \cline{2-4}
            & \ref{sec:niche-goals-social-language} & Social dynamics: language & learning by explanation \\ \cline{2-4}
            & \ref{sec:niche-goals-social-credit-blame} & Social dynamics: credit and blame & model others' intent; counterfactual reasoning \\ \hline
    \end{tabular}
    \caption{At-a-glance summary of ways that human causal cognition (right) is adapted to properties of the human niche (middle). Note that these are not one-to-one; many niche properties motivate the same cognitive capacity, or vice versa. Each row is elaborated in a section of text.}
    \label{tbl:niche-cognition}
\end{table}

\subsection{Human causal cognition is adapted to their environment}\label{sec:niche-environment}

\subsubsection{Humans have agency.}\label{sec:niche-environment-agency}

The statement that humans have agency is trivial but worth reflecting on. Acting in the world is the reason organisms have brains, process information, and form internal models in the first place \citep{fultotWhatAreNervous2019,robertsExamplesGibsonianAffordances2020}. In other words, \emph{doing} is always the ultimate goal of \emph{modeling}. From an agent's perspective, forming a causal model of the world is useful only insofar as it leads to useful actions and interventions in the future. As such, humans frequently learn and deploy incomplete and approximate causal models, but such models may nevertheless be useful to guide behavior \citep{bramleyConservativeForgetfulScholars2015,bramleyFormalizingNeurathShip2017}. In ML, model-free reinforcement learning is a prime example of how adaptive behaviors can emerge in a reactive system without any explicit internal model, let alone an approximate model \citep{suttonReinforcementLearningIntroduction2018}. Indeed, many organisms thrive in their respective niches with limited ability to form and reason with an internal model of the world. Humans, on the other hand, are generalist problem-solvers and thus benefit from planning in an internal model to guide actions, but that internal model is nonetheless an approximation to the true dynamics of the world.

Finally, having agency in the world means that, in many real-world cases, learning causal relations from observational data (i.e. in the absence of an intervention) is not necessary. Using one's agency to intervene on the world can dramatically simplify the causal learning problem, and this is indeed what humans do \citep{schulzLearningDoingIntervention2007}. While much of the causal learning literature asks the much more challenging question of how causality can be inferred from passive observational data, learning by intervening and observing the result simplifies the causal learning problem \citep{eberhardtInterventionsCausalInference2007,scherrerLearningNeuralCausal2021,ahujaInterventionalCausalRepresentation2023}.

\subsubsection{The environment is complex.}\label{sec:niche-environment-complex}

Agents living in the human niche face the daunting task of predicting and acting in an incredibly complex and interconnected world. Despite the enormous span of spatial and temporal scales and the sheer variety of things that one can potentially interact with, humans nonetheless develop and reason with predictive models that are generally correct, or at least \emph{useful}. One cannot hope to capture all of the interesting causal dynamics of the world with a single model. How, then, do humans do it? The human brain is very good at detecting patterns and regularities and at integrating that information into a vast knowledge base, and knowledge is not just of world dynamics but includes rich structures, schemas, concepts, and relations \citep{careyOriginConcepts2000,murphyBigBookConcepts2004,lakeBuildingMachinesThat2017}. One's understanding of the concept of a \emph{bicycle}, for instance, encompasses multiple causal and non-causal properties. Causally, a bicycle is a tool that affords transportation, it can be understood mechanistically in terms of wheels and gears, learning to ride a bike involves internalizing a dynamics model for balancing, etc. Further, this causal knowledge is linked to knowledge about other things, such as if knowledge of how chains and gears work transfers to better understanding the mechanisms of a clock. Just as important, however, are the non-causal properties: the concept of a bicycle also encompasses its typical appearance, how it is ontologically related to other forms of transportation, personal memories one has related to learning to ride a bike or past accidents and injuries, etc. Thus, one should think of causal descriptions in the human niche not as a model of the true or fundamental properties of the world, but as one component of a vast knowledge base about objects and experiences, ultimately aimed at organizing and making sense of a complex world. Objects have many properties; their causal properties are just one part of the story. 

The sheer complexity of the human niche --- and how humans nonetheless learn and deploy useful knowledge --- has multiple important implications for machine learning. First, contrary to the Reichenbach principle, not all relations are usefully modeled \emph{as causal relations}. Nor are all causal relations uni-directional. Instead, causality should be viewed as a subset of possible relations and as just one part of a broader knowledge base organizing the world into patterns and theories. As an example of this kind of thinking, note that the relation between concepts (``bicycle'') and properties (``has wheels'', ``can be used for transportation'') itself is a \emph{non-causal} kind of relation, which may explain why there is no consensus in the machine learning literature on whether, in classification tasks, categories cause observed data \citep{zhangDomainAdaptationTarget2013,gongDomainAdaptationConditional2016,lopez-pazDiscoveringCausalSignals2017a}, whether observed data cause categories \citep{rojas-carullaInvariantModelsCausal2018}, or some mix of these \citep{schoelkopfCausalAnticausalLearning2012,arjovskyInvariantRiskMinimization2019,waldCalibrationOutofdomainGeneralization2021}. 
Further, we argue that for complex domains like the human niche, emphasis should shift away from mathematical guarantees on identifiability and towards ``good enough'' and actionable models that strike a balance between simplicity, granularity, learnability, and efficacy on relevant tasks.

\subsubsection{The environment is open-ended.}\label{sec:niche-environment-open-ended}

What can an agent in the human niche do at any given time, and what observations might they make? The possibilities are practically unbounded. To say that the environment is \emph{open-ended} goes beyond saying that it is \emph{complex}. Closed systems can be complex, but open-ended systems allow for true novelty, creativity, and surprise \citep{stanleyWhyOpenEndednessMatters2019}. Open-endedness is a concept that originated in the fields of evolutionary algorithms and artificial life \citep{stanleyWhyOpenEndednessMatters2019}, and it has seen recent adoption in reinforcement learning \citep{ecoffetFirstReturnThen2021} and robotics \citep{cullyRobotsThatCan2015}. \citet{ibelingOpenUniverseCausalReasoning2020} provide a formal, axiomatic treatment of the problem of causal reasoning with an unknown number of variables, paving the way for further formal work on open-endedness.

Humans are adapted to an open-ended world in two ways. First, humans face the challenge of open-endedness through curiosity and hypothesis-driven exploration of new possibilities.
Hypothesis-driven exploration is the strategy of first hypothesizing options, then taking actions that are most likely reveal information that distinguishes those hypotheses. The scientific method itself is an example: where humans lack the ability to directly measure something, they devise new measurement tools, form hypotheses (often as causal models), and test those hypotheses on data. It is well-documented that children engage in goal-directed and hypothesis-driven exploration as a form of play \citep{careyConceptualChangeChildhood1987,gopnikWordsThoughtsTheories1997}. ``Theory-Based Reinforcement Learning'' implements these ideas in the framework of model-based reinforcement learning, with promising results \citep{tsividisHumanLevelReinforcementLearning2021}. The related ideas of active inference and epistemic foraging formalize curiosity as an active search for information about the world \citep{fristonActiveInferenceCuriosity2017,parrUncertaintyEpistemicsActive2017}.

Second, humans face the challenge of open-endedness by through causal induction, or the ability to instantiate new causal models on the fly in novel situations and to adapt the scope or level of detail of the simulation to the problem at hand \citep{waldmannKnowledgeBasedCausalInduction1996,gopnikWordsThoughtsTheories1997,griffithsTheorybasedCausalInduction2009,tenenbaumIntuitiveTheoriesGrammars2007,griffithsTwoProposalsCausal2007,griffithsFormalizingPriorKnowledge2017}. Causal induction underlies the remarkable fact that humans make effective causal predictions in novel situations where they have no explicit prior experience. Consider this example: ``what would happen if a balloon fell onto the back of a porcupine?'' One might induce, ``the balloon would pop, the porcupine would be startled.'' Causal induction requires organizing causal knowledge at a higher level of abstraction, learning \emph{models of causal models}, also known as ``causal theories'' or ``causal overhypotheses.'' Whereas a particular causal model applies to one particular environment or situation, a causal theory instantiates new causal models on the fly in novel situations, e.g. generalizing from experience or knowledge that needles pop balloons and recognizing that porcupine quills would likely cause a similar outcome. In ML terms, this is ``zero-shot'' learning of causal models in novel situations. Given a causal theory, a causal model might be instantiated only once in a novel situation, but may nonetheless make accurate zero-shot predictions and provide useful counterfactual explanations for a single outcome in a unique setting. Effective causal theories build on an ontological understanding of objects and their properties \citep{nilforoshanZeroshotCausalLearning2023,kanskySchemaNetworksZeroshot2017,nairCausalInductionVisual2019a}, they can be deployed at various levels of granularity depending on the demands of a given situation \citep{gopnikWordsThoughtsTheories1997,griffithsTheorybasedCausalInduction2009}, and they may instantiate a model that relates only the immediately relevant variables in the present situation while leaving irrelevant details unspecified. These considerations of ontology, granularity, and relevant/irrelevant details are discussed further in other sections.

\subsubsection{The environment is almost always confounded.}\label{sec:niche-environment-confounded}

Agents in the human niche cannot hope to measure everything in their environment. In almost all aspects of life, there is the potential for hidden causes that could explain observed correlations. In other words, any observed relations in the world are potentially confounded due to hidden common causes.\footnote{Other implications of limited observation are discussed in Section \ref{sec:niche-constraints} below, but the issue of confounding is fundamental to all agents in the human niche, so we are organizing it as a property of the environment here.} 

The fundamental challenge of confounding is easy to show in a simple linear regression setting. Assume a pair of observed $\vx$ and $\vy$ are affected by shared unobserved $\vz$: $p(\vy|\vx) = \int p(\vy|\vx,\vz)p(\vz)d\vz$. The introduction of such unobserved confounders makes it impossible to accurately estimate the causal relation between $\vx$ and $\vy$ based only on observations. For example, in linear confounding, we have $\vy = \beta \vx + \delta \vz + \noise$, where $\beta$ and $\delta$ characterize the causal influences of $\vx$ and $\vz$ on $\vy$ and $\noise$ is noise. Linear regression of $\vy$ on $\vx$ then does not give us the causal influence of $\vx$ on $\vy$ but rather gives us a biased estimate:
$$\mathbb{E}[\beta^*] = \beta + \delta (\mX^\top \mX)^{-1}\mathbb{E}[\mX^\top \mZ | \mX]$$
(this is the well-known omitted-variable bias). This simple example shows that biases can become arbitrarily large as the world becomes strongly confounded.

To address the ever-present possibility of unobserved confounders, agents in the human niche require epistemic humility --- an understanding that one's best model of the world is always subject to revision in light of new information. One approach is to maintain not a single model but a distribution over plausible causal models (see Sec. \ref{sec:niche-constraints-observations} below), and there is some evidence that humans do this consistent with Bayesian inference over models \citep{tenenbaumHowGrowMind2011}. However, this is only a partial solution because unobserved confounders cannot be instantiated in a model without other strong assumptions \citep{wangBlessingsMultipleCauses2019}.

Humans and other agents in the human niche have a special power to overcome confounding: they can test their theories by running experiments on the world. One does not merely observe a correlation between the state of a light switch and the state of a light --- a skeptic would point out that this could be confounded by the actions of others who may be controlling both the switch and the light. This ambiguity disappears when you toggle the switch yourself and the light turns on an instant later. Skepticism about confounders is addressed by randomizing one's own actions. In the case of the light switch, the precise time that you flip the switch is effectively random, ruling out other potential explanations such as the light being on a timer. A skeptic may still worry that they are in an adversarial environment or that their own preferences and actions are manipulated by the environment in some unknown way. This concern is mathematically warranted \citep{bareinboimBanditsUnobservedConfounders2015}, yet it nonetheless seems reasonable for an agent in the human niche to assume that such adversaries are rare.

Epistemic humility goes hand-in-hand with curiosity, experimentation, and hypothesis-driven exploration. An experiment by \citeauthor{schulzSeriousFunPreschoolers2007} showed that preschoolers were sensitive to confounding and played more with toys when they were less certain of the toy's function, consistent with a theory in developmental psychology that children's play is an act of exploration and hypothesis-testing to learn about the world \citep{gopnikScientistChild1996,gopnikReconstructingConstructivismCausal2012}.

\subsubsection{The environment is designed by humans.}\label{sec:niche-environment-designed}

Much of the human niche is designed and built by humans and for the benefit of humans. Better appreciating how humans conceptualize causality thus provides an additional lens both on how the environment itself is structured, as well as on how to design an intuitive environment. For example, humans have an inductive bias of preferring simple and sparse causal explanations, and so we design objects with causal power (like a lightswitch) to have a simple and sparse interface that supports atomic interventions and well-controlled effects. In this way, humans have the power to match the environment to their simplifying assumptions, rather than the other way around. It follows that ML systems operating in the human-built environment have an even stronger incentive to include human-like causal inductive biases to facilitate learning about human-created systems. These inductive biases become doubly effective in environments that are designed to be intuitive for humans.

\subsubsection{Other agents exist.}\label{sec:niche-environment-others}

The human niche involves interacting with other --- both human and non-human --- agents with their own goals, constraints, and ability to intervene on the world. Interactions with other agents can be cooperative or competitive, and humans readily infer others' goals and intents from their actions \citep{gopnikWordsThoughtsTheories1997}. They then use that information to learn about causal properties of the world by observing others' actions and demonstrated interventions. Some of these abilities emerge at remarkably young ages: infants as young as 6 months old distinguish between the motion of inanimate objects and the intentional actions of other humans \citep{woodwardInfantsSelectivelyEncode1998}. Implementing the capacity to learn from others in ML systems can build on computational models of action understanding \citep{bakerActionUnderstandingInverse2009}, inverse reinforcement learning \citep{ngAlgorithmsInverseReinforcement2000}, and causal learning in the face of unknown interventions \citep{keLearningNeuralCausal2020}. However, further bridging the gap to human cognition may require a richer language for expressing goals, intentions, and desires than can be expressed in a reward function \citep{abelExpressivityMarkovReward2021,Velez-ginorio2016,davidsonGoalsRewardProducingPrograms2024}.

\subsubsection{The environment has hierarchical structure.}\label{sec:niche-environment-hierarchical}

The complexities of the world make high-fidelity modeling challenging, but its complexity is not without useful and exploitable structure in both space and time. Spatially, objects can be understood in terms of their parts or grouped together as a collection. Consider, for example, modeling causal interactions in collisions between marbles: a one-on-one collision between two marbles presents a straightforward causal relationship that is well understood by anyone who has played a game of billiards. However, when confronted with a multitude of marbles spilling across a surface, tracking individual collisions becomes infeasible. Nonetheless, humans intuitively form predictions about the aggregate statistics of complex scenes or objects \citep{battagliaSimulationEnginePhysical2013}. In such instances, a higher-level, or coarse-grained, approach is adopted, summarizing the system through aggregate statistics like mean velocity or total volume of the pile of marbles. 

Temporally, detailed causal relations that unfold over short timescales often have structure that can be abstracted at longer timescales. As in the example of a spilling pile of marbles, individual details are lost in aggregating or coarse-graining in time, but coarse-grained statistics --- over long timescales from days to decades --- may nonetheless be predictable. Consider the problem of choosing where to go for lunch. This planning problem can be abstracted to a high level (where to go, what route to take) and ignore low-level details (particular muscle movements of each individual step along the way). Humans naturally make causal predictions at a range of spatial and temporal scales, often choosing the scale that is adaptive to the task at hand.

These examples highlight the human tendency to abstract away from the minutiae of complex phenomena, summarizing and abstracting in a way that keeps the number of interacting parts manageable. This reflects, in part, an adaptation of a finite and constrained brain to a complex yet hierarchically structured world: if there is a maximum number of things that you can attend to and process at once, then it becomes necessary to lump objects or events together so as to reduce the number of discrete entities being modeled. Further, it is another good reminder that \emph{good enough} causal predictions are those that are both computationally feasible and guide successful behavior, not necessarily those that are detailed and veridical. Many complex systems are challenging to model with high fidelity even with powerful computing resources, and so the human tendency to model coarse-grained causal relations is a useful adaptation for any constrained computing system modeling a complex world, not only in biological agents but in any bounded-rational system \citep{Genewein2015}.

Humans do not just select one scale at which to model a system, but many scales \citep{slomanCasualModelsHow2009}. Planning where to go for lunch on a given day may fit into a longer-timescale plan for dieting and exercise, and it may depend on shorter-timescale nuances like whether one has comfortable footwear for a long walk. The aggregate statistical behavior of a pile-of-marbles can figure into one's understanding of an hourglass (by analogy between marbles and grains of sand), or the pile may be understood or explained in terms of the detailed collisions among its constituents. This multi-scale understanding of systems is key to how humans make sense of and explain causal systems: \emph{causal mechanisms} in human explanations often themselves consist of one-level-down causal systems with their own mechanistic explanations, etc. The mechanics of riding a bicycle are explained (in part) in terms of gears and gear ratios. Gears and chains are explained in terms of sprockets and chain links pushing and pulling on each other. The physical interactions are explained in terms of material or atomic properties of the metal, etc. Thus, understanding the causal interactions at one level of a system both informs and is informed by the causal interactions at other levels of abstraction. Building machine learning systems that are capable of such multi-level modeling where each level constrains or informs the others is a promising direction for future work.

Coarse-graining has recently received attention in machine learning, but it remains a largely open and unsolved problem to precisely characterize the costs and benefits of hierarchical or multi-scale models. \citeauthor{rubensteinCausalConsistencyStructural2017} identify casual self-consistency relations that should hold across scales. An interesting extension is to \emph{approximate} coarse-graining where some violations of self-consistency across scales are acceptable if they make the model sufficiently simple \citep{beckersApproximateCausalAbstraction2019}. An interesting direction for future theoretical work will be to optimize the trade-off between a simplicity and veridicality of an agent's causal model(s), taking into account the agent's goals and constraints: the best causal model is one that captures relevant aspects of the environment well enough while being simple and efficient simulate \citep{Genewein2015}.

\subsubsection{The environment has ontological structure.}\label{sec:niche-environment-ontology}

Ontological structure is how we organize our knowledge of things in the world based on similarities and differences among their properties. Cups, mugs, and bowls are all related in virtue of their shape and the property that they can hold liquid. A colander shares some properties with a bowl (shape, holds a collection of things above a certain size) and others with a net (allows liquid or small enough things to pass through). Understanding objects' properties and how they relate to other objects is a critical ingredient in how humans organize and generalize their causal (and non-causal) knowledge about the world \citep{holyoakInferringCausalRelations2017}.

It is a slight abuse of the term to say that ontological structure is a property of the environment; it would be more accurate to say that an ontology is in the mind of the beholder. However, to answer the question of \emph{why} organizing knowledge into properties and ontologies is useful for an agent in the human niche, one can begin by appreciating that things in the world \emph{actually do} share common properties, independent of how any particular observer organizes them. The laws of physics apply to all objects. Things in the natural world share common properties with each other due to shared evolutionary history. Things in the human-built environment share properties with each other due to shared technological history.

Recognizing similarities and differences among objects (or people, events, actions, etc) in this way enables humans to build a rich and interconnected internal model of the world and to form causal analogies and metaphors \citep{holyoakInferringCausalRelations2017}. Rather, it is more accurate to say that humans do not have ``an'' internal model, but many causal models for different domains and situations, instantiating new causal models on the fly. Ontology is therefore a critical component of causal induction and causal theories (Sec. \ref{sec:niche-environment-open-ended}).

\subsubsection{Interactions are typically sparse.}\label{sec:niche-environment-sparse}

While the world is complex and full of many objects and agents that could \emph{potentially} interact, in practice interactions are sparse in both space and time. In an average everyday scene, most objects are stationary, interactions between agents and objects are rare (at least in terms of interactions that are relevant to a given agent), and when interactions do occur, they tend to involve a small number of objects or actors and affect only a subset of their properties. All of these kinds of sparsity can be and have been leveraged as inductive biases to simplify causal learning and causal modeling.

Humans take advantage of the sparsity of causal events to simplify how they model the world. The way humans learn and reason about causality is mediated by the limited capacity of attention and working memory, placing the ``spotlight'' on a few entities and their interactions at a time. Simplifying causal explanations into few interacting entities is a known bias in humans' causal judgments \citep{lombrozoSimplicityProbabilityCausal2007}. This bias towards reasoning about sparse --- often dyadic --- causal interactions may be a feature, not a bug; it reflects an adaptation of applying limited computational resources (see Section \ref{sec:niche-constraints-resources}) to modeling a world where relevant causal interactions really are typically sparse. Occam's razor may not always lead to true conclusions, but it is nonetheless a useful heuristic.

Several methods for incorporating sparsity assumptions into causal models have already been developed in machine learning. First, as discussed in section \ref{sec:scm_model_structure}, work within the SCM framework often exploits an assumption about the maximum degree or degree distribution of a causal graph to facilitate learning \citep{claassenLearningSparseCausal2013}. However, this approach typically still assumes that the goal of causal learning is to describe a dataset or domain in terms of a single (sparse) directed acyclic graph. Another framework incorporating sparsity is so-called ``object-centric'' methods that seek to learn causal models by explicitly representing objects as distinct entities and including an inductive bias that interactions between objects are sparse \citep{elsayedSAViEndtoEndObjectCentric2022,wuSlotFormerUnsupervisedVisual2022}. Still other methods have incorporated sparse changes over time as an inductive bias in representation learning \citep{klindtNonlinearDisentanglementNatural2021}. Inspired by human cognition, we advocate for further work combining sparse causal interactions, attention or capacity constraints, explicit object representations, and multi-level coarse-grained entities.

Newtonian physics provides some challenges for causal models. Consider the case of a moving ball that collides with and ``launches'' a stationary ball. Human intuition would say that the moving ball \emph{caused} the stationary ball to start moving \citep{michottePerceptionCausality2017}. However, it is equally true that the stationary ball \emph{caused} the moving ball to stop or change direction. In fact, viewing the same scene from a moving reference frame, it may appear that the first ball is stationary, and the second ball moves into the scene and launches the first. Physically, there is an equal and opposite interaction between the two billiard balls which is interpreted in a causally asymmetric way by human observers. This kind of interaction can, in principle, be captured by an SCM that ``unrolls'' the physical dynamics over time. Although \emph{physically accurate}, such a model is cumbersome to use and does not capture the clear asymmetry in humans' intuitive judgements, which depend on the reference frame to determine which ball ``launches'' the other \citep{chengCausalInvarianceEssential2017}. This asymmetry in human judgements of symmetric events may be the result of a sparsity prior: while the Newtonian physics of collisions are independent of an observer's (constant velocity) reference frame, sparsity in the sense that \emph{most} objects are stationary most of the time suggests that there is a privileged reference frame in the world. Thus, humans attribute agency and more causal power towards things that move relative to this reference frame than things that are stationary.

\subsubsection{Stream of data.}\label{sec:niche-environment-stream}

Agents in the human niche experience the world as a continuous stream of observations and actions. Further, the world itself is constantly in flux; an open-ended world generates a continuous stream of surprises. As the saying goes, \begin{quote}``No man ever steps in the same river twice, for it's not the same river and he's not the same man.''\footnote{Attributed to the ancient Greek philosopher Heraclitus: \url{https://plato.stanford.edu/entries/heraclitus/}}\end{quote} Thus, one must extract general causal principles because it cannot be assumed that past states will exactly recur. Causal learning in the human niche is a matter of lifelong learning in an ever changing environment \citep{silvaProspectiveLearningPrincipled2023}. It is also a matter of lifelong model deployment; states never exactly repeat and there are no definitive episode boundaries, but humans nonetheless make generally accurate causal predictions in novel situations. This is again a feature of causal induction and causal analogy-making (see \ref{sec:niche-environment-open-ended} and \ref{sec:niche-environment-ontology} above).

Humans also segment their stream of experiences and store them in episodic memory \citep{ezzyatWhatConstitutesEpisode2011}. Episodic memory is then revisited or elaborated on counterfactually to aid learning and generalization to new situations that are similar to those encountered in the past. Reasoning counterfactually about past events has been shown to make learning more efficient in reinforcement learning settings \citep{buesingWouldaCouldaShoulda2018,harutyunyan_hindsight_2019}.

\subsubsection{Arrow of time.}\label{sec:niche-environment-arrow-of-time}

Causes precede their effects, and effects often follow immediately after their causes. Time thus gives two important cues to causality: order and simultaneity. For agents who experience a continuous stream of inputs, time provides a powerful learning signal \citep{bramleyTimeCausalStructure2018}. Of course, not all simultaneous events are causally related, and so a learner who assumes this as an inductive bias will make occasional mistakes. Human cognition, with its remarkable capacity for pattern recognition, sometimes leads us to perceive causality between unrelated but simultaneous events. For instance, if we hear a loud sound the moment a light clicks off, we may momentarily attribute the sound as the cause of the lights turning off \citep{siegelVisualExperienceCausation2011}. This is perhaps a kind of ``causal illusion'' \citep{gopnikTheoryCausalLearning2004} that reveals a strong inductive bias where humans use simultaneity as a strong cue for causality.

An explicit account of time is typically absent from the SCM framework. Instead, the SCM framework emphasizes steady-state or equilibrium relations between causally related variables \citep{scholkopfCausalityMachineLearning2019}. To represent time explicitly within an SCM, one might resort to "unrolling" dynamics, wherein states at time $t$ become causal parents of states at time $t+1$ (as illustrated in the billiard ball example earlier). However, this approach can be cumbersome and fails to provide useful abstractions beyond a mere representation of dynamical systems. In contrast, humans reason causally at a variety of different timescales, adapting to the task at hand. For example, one may reason on a short timescale about how knocking over a single domino will affect the next domino in the chain, or one may reason over a longer timescale about how the entire chain of dominoes will eventually --- through some intuitive prediction --- end up in the fallen-over state \citep{jayaramanTimeagnosticPredictionPredicting2019}. The related idea of coarse-graining time has a long history in AI \citep{sacerdotiPlanningHierarchyAbstraction1974,suttonMDPsSemiMDPsFramework1999}, 

Time plays an important but implicit role in the SCM framework in the sense that it is tied to the definitions and assumptions of the framework \citep{petersElementsCausalInference2017}. For instance, the arrow of time itself, the asymmetry of causality, and the statistical (algorithmic) independence of mechanisms from initial conditions are all inextricably intertwined \citep{janzingAlgorithmicIndependenceInitial2016}. Thus, certain assumptions of the SCM framework, such as asymmetric causal relations, are deeply rooted in time and the idea that causes must precede their effects. However, as already discussed, this is a rather different approach to time than explicitly modeling temporal dynamics under interventions, which is necessary for multi-step planning and appears to be an integral part of how humans form mental models of the world.

Time also plays an interesting role in quasi-experimental methods for inferring causality from observational data. Some quasi-experimental designs such as regression discontinuity or differences-in-differences leverage naturally occurring variations over time in so-called \emph{instrumental variables} to infer causality \citep{liuQuantifyingCausalityData2021}. These designs often rely on the assumption of temporal precedence, where the cause precedes the effect, to establish causal relationships. Understanding how humans navigate and interpret such temporal cues could provide valuable insights for designing machine learning algorithms capable of capturing causality in observational data.

\subsection{Human causal cognition is adapted to their constraints}\label{sec:niche-constraints}

\subsubsection{Observations are partial.}\label{sec:niche-constraints-observations}

Operating in the human niche requires the ability to cope with missing, noisy, or ambiguous data as well as small sample sizes. Within the SCM framework, missing observations (hidden variables) may be considered ``noise'' if they are ancestors of only one endogenous variable, and can be ignored entirely if they are descendants of all endogenous variables. Missing data or missing observations for variables that are ancestors of two or more endogenous variables are considered \emph{hidden confounders} and are much more problematic. By and large, the most common way of dealing with hidden confounders in the SCM framework is to \emph{assume they do not exist} \citep{petersElementsCausalInference2017} or to assume some slightly weaker conditions on their structure \citep{wangBlessingsMultipleCauses2019}. These assumptions not only restrict the scope of the SCM framework, but they also place high demands on observation itself: one must be diligent not to miss any observations and ensure data are clean. While the no-unobserved-confounders assumption is valuable and often applicable in the tabular setting,\footnote{By ``tabular setting,'' we mean problem domains where data are collected either by specialized measurement tools or by careful data-entry so that a discrete set of pre-specified variables are columns and observations are rows of a table.} it is too strict for generalist agents in the human niche who observe the world as a continuous data stream and who must accept necessary limitations on observations.

Humans cope with missing or ambiguous information in three ways:

\begin{enumerate}
    \item \textbf{Simplify.} Applying Occam's Razor, one may conclude that, all else being equal, causal relations likely involve few interacting entities, brief intervals, and influence only a few properties of those entities. This is related to the various kinds of \emph{sparsity} discussed above (Sec \ref{sec:niche-environment-sparse}). Faced with uncertain or ambiguous data, applying Occam's Razor to select simple causal explanations over more complex explanations is only a heuristic, but it can be useful.
    
    \item \textbf{Represent uncertainty.} Limited observations and limited ability to intervene entail limited information about causal structures (causal models or causal theories) themselves, and so agents in the human niche can benefit from maintaining uncertainty in the form of a distribution over plausible causal models (or theories) that are consistent with observations so far. There is some evidence that humans reason probabilistically over causal models in this way, a key indicator being that different causal structures can explain each other away in light of new data. For example, \citeauthor{vasilyevaDevelopmentStructuralThinking2018} showed that both children and (to a greater extent) adults will attribute another person's observed behavior to external rather than internal factors --- providing an alternative causal explanation --- in light of some indirect evidence that external factors may have been at play \citep{vasilyevaDevelopmentStructuralThinking2018}. Such results suggest that both children and (to a greater extent) adults maintain, perhaps implicitly, a set of plausible causal explanations, and revise their beliefs about those explanations in light of new evidence.
    
    Formally, representing uncertainty over models involves the posterior distribution $p(M|D)$ of possible causal models $M$ given data $D$. The prior distribution over models $p(M)$ can express a large variety of constraints or inductive biases such as preferring sparse causal models to dense ones. Priors over certain types of causal models are also a way to express causal theories \citep{Kemp2007,griffithsFormalizingPriorKnowledge2017}. Ideally, causal predictions would then be averaged over a posterior predictive distribution, averaging together predictions from many models while weighting them by the models' posterior probability. However, representing a full posterior distribution over possible models is intractable due to the combinatorially large spaces involved, e.g. requiring time exponential in $n$ simply to compute marginal probabilities of edges in a DAG with $n$ variables \citep{koivistoAdvancesExactBayesian2012}. One type of solution to this problem proposed in the cognitive science literature is to represent only a small set of plausible hypotheses at a time, analogous to sampling or filtering methods \citep{denisonRationalVariabilityChildrens2013,bramleyFormalizingNeurathShip2017}. More expressive and more powerful methods for representing posteriors over models are available to AI/ML systems. For instance, \citeauthor{deleuBayesianStructureLearning2022} used Generative Flow Networks or GFlowNets \citep{bengioGFlowNetFoundations2022} to amortize the model inference process and efficiently sample from the posterior over DAGs.

    \item \textbf{Curiosity, active inference, and hypothesis-driven exploration.} When missing some critical information, one can sometimes take actions that reveal the missing information. This process goes by different names: active inference, hypothesis-driven exploration, and curiosity. Curiosity benefits from the previous bullet point of representing uncertainty. ``Knowing what you don't know'' can lead to more targeted interventions, taking actions that maximize the expected reduction in uncertainty both over the current state of the world (active inference) and over underlying structures (hypothesis-driven exploration) \citep{fristonActiveInferenceCuriosity2017}. Developmental psychology has shown that children actively explore their environment to test out different hypotheses \citep{gopnikScientistChild1996,schulzSeriousFunPreschoolers2007}. The results of \citeauthor{kosoyLearningCausalOverhypotheses2022} suggest that young children between 4 and 6 years old take actions to near-optimally disambiguate possible causal models of a toy, and further identify the problem of \emph{expressing a sufficiently rich class of hypotheses about causal models} (or ``overhypotheses'') as a key gap between current artificial systems and human children and an open challenge for reinforcement learning \citep{kosoyLearningCausalOverhypotheses2022}. \citeauthor{poliCuriosityDynamicsOptimal2024} provide a recent synthesis of computational models of curiosity in developmental psychology and cognitive science \citep{poliCuriosityDynamicsOptimal2024}.

    Computational models of curiosity have proliferated in reinforcement learning and robotics, especially in model-based settings \citep{schmidhuberPossibilityImplementingCuriosity1991a,oudeyerWhatIntrinsicMotivation2007,bartoIntrinsicMotivationReinforcement2013}. Past work has shown repeatedly that reinforcement learning agents can benefit significantly from intrinsic motivation, or internally-generated rewards that encourage useful kinds of exploration. One source of intrinsic motivation is prediction errors about the next observation or latent state in a sequence \citep{schmidhuberPossibilityImplementingCuriosity1991a,ororbiaActivePredictiveCoding2023,mazzagliaCuriosityDrivenExplorationLatent2022}, encouraging an agent to seek out (slightly) unpredictable states. Classically, dynamics models in reinforcement learning and control are not discussed in the language of causality, despite being models that inherently make predictions about future states conditioned on actions (interventions). In the absence of an explicitly causal world model, it can be awkward to distinguish between the autonomous dynamics of other objects and entities in the world and dynamics which the agent causally influences. Further building towards explicitly causal world-models, in conjunction with curiosity and hypothesis-driven exploration, is a promising area of ongoing research.
\end{enumerate}

\subsubsection{Interventions are sometimes infeasible.}\label{sec:niche-constraints-infeasible-intervention}

While learning-by-doing is simpler than learning from observational data, agents in the human niche must accept that not all interventions are feasible, whether due to physical limits, ethical concerns, or other costs. This is where causal induction with a causal theory and a mechanistic understanding of typical causal relations can be especially useful: taking insights from related domains and making causal analogies can inform decision-making even where direct experimentation is infeasible. For example, if one has had experience with fire being hot and causing pain, and one knows that a stove works by getting hot, one does not need to touch a hot stove to learn that it, too, causes pain. This example illustrates that mental simulation can still be accurate and useful even for predictions where one does not have direct experience. 
\subsubsection{Limited attention and working memory.}\label{sec:niche-constraints-resources}

All agents in the human niche have limited capacity to process available information in the world, and this is not something to be solved by scaling up agents. Relative to the sheer complexity and open-endedness of the environment, all reasonably efficient agents must ignore some of the information coming in. Further, humans have a finite capacity for working memory. How do humans excel at causal reasoning about a complex and open-ended world while constrained to using finite resources?

\begin{enumerate}
    \item Humans compress complex scenes with numerous complex entities into smaller sets of coarse-grained entities and temporally-extended events; see Sec. \ref{sec:niche-environment-hierarchical} above. Coarse-graining can thus be seen as an adaptation simultaneously to the hierarchical complexity of the environment itself as well as the limitations of finite agents reasoning about a complex world.
    \item Humans focus on small numbers of objects or entities at a time, often attending to dyadic relations between just two (possibly coarse-grained) objects; see Sec. \ref{sec:niche-environment-sparse} above.
    \item Humans focus on the relevant properties of those objects or entities as demanded by the present situation. For example, consider the various kinds of causal stories an object like a \emph{bicycle} can participate in. When planning a commute, one can focus on a bicycle's properties as a vehicle, ignoring its mechanistic details. When repairing a broken chain, one can then focus on the mechanistic details, and one can even apply that knowledge to novel situations such as repairing a friend's fixed-gear bicycle. In each situation, relevant causal information about the bicycle is retrieved and a model is instantiated on the fly at an appropriate level of granularity. One does not need to call to mind the details of how chains and sprockets work in order to plan a route.
\end{enumerate}

\subsubsection{Reasoning is slow.}\label{sec:niche-constraints-reasoning-slow}

\begin{quote}
    ``Encoded in the large, highly evolved sensory and motor portions of the human brain is a billion years of experience about the nature of the world and how to survive in it. The deliberate process we call reasoning is, I believe, the thinnest veneer of human thought, effective only because it is supported by this much older and much more powerful, though usually unconscious, sensorimotor knowledge. We are all prodigious olympians in perceptual and motor areas, so good that we make the difficult look easy. Abstract thought, though, is a new trick, perhaps less than 100 thousand years old. We have not yet mastered it. It is not all that intrinsically difficult; it just seems so when we do it.'' \citep{moravecMindChildrenFuture1995}
\end{quote}

In the language of System-1 and System-2 cognition \citep{kahnemanThinkingFastSlow2011}, this quote by Moravec highlights how fast the automatic pattern-recognition capabilities of System-1 seem and how slow the deliberate serial reasoning of System-2 seems for humans, but also much of this discrepancy is explained by the fact that the System-2 capabilities of the human brain are a recent development in evolutionary terms.

Humans overcome some of the slowness of System-2 reasoning by amortizing at all levels of causal reasoning. Amortization in this context is the process of learning System-1 shortcuts --- fast pattern-recognition --- that approximates what would be the outcome of slower deliberative System-2 reasoning. In general, inference problems can be amortized by treating inference as a function approximation problem and training a model to produce approximate inferences when fed some observation or data \citep{stuhlmuellerLearningStochasticInverses2013}. Multiple theories in neuroscience and cognitive science involve amortizing slow recurrent processes \citep{dayan_recognition_1997,gershmanAmortizedInferenceProbabilistic2014,dasguptaTheoryLearningInfer2020}. In the context of causality, amortization can be applied at many levels. First, amortized inference can be applied to rapidly infer latent variables conditioned on observations through the use of so-called recognition models \citep{dayan_recognition_1997,kingmaAutoencodingVariationalBayes2014}. Second, at one level more abstract, amortized inference can be applied to rapidly infer model structure from an entire dataset \citep{keAmortizedLearningNeural2020,deleuBayesianStructureLearning2022}. An interesting direction for research will be to take this one step more abstract and amortize the learning of causal theories themselves.

\subsubsection{All models are wrong.}\label{sec:niche-constraints-models-wrong}

Forming a true and complete causal model of the world is not achievable. Yet, as discussed in Section \ref{sec:niche-environment-agency}, agents in the human niche benefit from forming predictive models of the world that are \emph{useful} in the sense of guiding their actions. Let us expand here on different ways that a causal model can be ``useful'':

\begin{enumerate}
    \item Useful models include those that make precise and accurate predictions on future data. Importantly, predictive power does not necessarily require a detailed mechanistic understanding. A classic example is the ideal gas law, or $PV=nRT$, which has high predictive power for how a gas will behave in various settings --- both observational and intervening on $P$, $V$, or $T$ --- despite leaving underlying mechanisms unspecified. Another example is the folk wisdom that a red sky at night forecasts good weather while a red sky in the morning forecasts bad weather. This is a useful predictive model that dates back millennia, but which again leaves the mechanistic details for why it is true unspecified.
    
    \item Useful models include those which provide insight into underlying mechanisms. Take Newton's theory of gravity and Freud's psychoanalytic theory as examples. Both are scientific theories which seek to explain some phenomenon (gravitation, human behavior) in terms of some underlying mechanistic process (gravitational force that obeys an inverse square law, competition among the id/ego/superego). Both are also considered to be disproven, yet each is taught to new students entering physics or psychology. The utility of these models goes beyond their simplicity and predictive power: Newtonian physics teaches students to think in terms of objects obeying physical laws, and Freudian psychoanalytic theory can in some cases lead to useful therapeutic interventions.
    
    Understanding mechanisms --- even incorrect mechanisms --- can also be useful for a generalist agent building a causal theory about the world. Planets may not \emph{actually} attract each other by Newtonian gravitational forces, but understanding forces of attraction and how they give rise to orbits is itself a concept that generalizes. Mechanistic models can thus also give rise to useful causal analogies in novel domains.
    
    \item Useful models include those that balance moderate errors with simplicity and ease-of-use. A model must, by definition, provide some abstraction. The simpler and easier-to-use a model is, the more its inaccuracies can be tolerated. Humans often get by with simple heuristics rather than detailed internal models in situations where the added computational burden of a higher-fidelity model is simply not worth it.
\end{enumerate}

It is well-known among statisticians that predictive uncertainty (item 1) and inferential uncertainty (item 2) are distinct in the sense that, when fitting some model to data, low inferential uncertainty does not guarantee low predictive uncertainty, and vice versa. This insight bears on the identifiability of a causal model from data. Essentially, a model is identifiable if inferential uncertainty approaches zero in the limit of infinite data. However, low inferential uncertainty alone cannot guarantee low predictive uncertainty nor simplicity of the model. There can be non-identifiable situations which nonetheless entail high predictive accuracy, and there can be identifiable situations which provide mechanistic insight but no predictive power. Finally, both of these situations can happen with simple or highly complex models. Thus, ML may benefit from a theory of causal learning that balances all three criteria of prediction, inference, and simplicity, and that moves away from model identifiability as the sole target of causal inference.

Finally, having a wrong model may motivate humans to reason counterfactually; events may be difficult to predict, but relatively easy to explain in hindsight \citep{byrneCounterfactualThought2016}. Constructing explanations of past events in terms of plausible counterfactual alternatives is an effective way to learn in the absence of an accurate predictive model \citep{abbeelUsingInaccurateModels2006,buesingWouldaCouldaShoulda2018,jannerWhenTrustYour2019}.

\subsection{Human causal cognition is adapted to their goals}\label{sec:niche-goals}

\subsubsection{Goals are compositional and coarse-grained.}\label{sec:niche-goals-compositional}

The structure of any agent's goals informs how they model and interact with their environment. While humans operate in what can be viewed as extremely high-dimensional observation- and state-spaces with complex fine-scale dynamics, the most relevant aspects of the environment --- those things that enter into humans' reward functions, so to speak --- are themselves typically abstract or coarse-grained states. The goal of satisfying hunger, for example, is achieved by eating some food, but to a large extent it does not matter what food is eaten, how, or where. Reasoning about this goal and planning towards it is likewise naturally done in a coarse-grained model.

Humans' goals are not just coarse-grained, but also compositional, meaning that they can be expressed in terms of sub-goals and relations among entities. One way to achieve the goal of not being hungry is to cook, which depends on the sub-goal of acquiring ingredients, and the order in which those ingredients are acquired does not matter. To capture this rich structure, \citeauthor{Velez-ginorio2016} proposed expressing humans' goals in the form of compositional programs over abstract states, and found that this model captures much of how humans infer each others' goals from their actions \citep{Velez-ginorio2016}. Thus, in the context of inverse reinforcement learning or learning from demonstration, reward functions should be expressed compositionally in terms of abstract states and their relations to better match the structure of goals which humans are demonstrating.

Expressing an agent's goals in compositional terms thus benefits humans who work with those agents, but further this compositionality may directly benefit the agent. In an open-ended environment (Section \ref{sec:niche-environment-open-ended}), agents can benefit from ongoing exploration, curiosity, and hypothesis testing. Hypotheses about how the world works are themselves naturally compositional; for instance, a child who learns that pressing a red button causes a red light to turn on may be curious about a green button and hypothesize that it will turn the light green. Along these lines, \citeauthor{colasLanguageCognitiveTool2020} recently proposed using language-like compositionality to drive hypothesis-driven exploration in an artificial agent and found it to be an effective method to explore an open-ended environment.

\subsubsection{Goals are contextual and change over time.}\label{sec:niche-goals-contextual}

For a variety of reasons, agents in the human niche must be prepared to adapt their goals on the fly. One reason is that once a goal is achieved, it may no longer be a goal. For example, finding food may be the goal while hungry, but once fed, one's goals shift to other things. Another reason why goals change is if the environment changes or presents an unexpected obstacle. For example, imagine riding a bicycle with the goal of meeting up with a friend. If the bicycle gets a flat tire along the way, one's immediate goal may shift to fixing the bicycle, which is now seen as a necessary intermediate goal towards reaching the original destination. 

The world is highly complex, and agents in the human niche benefit from being selective about the kinds of information they process. When goals change, different aspects of the environment become relevant. In the example of a bicycle with a flat tire, one's conception of the bicycle itself changes when the goal changes. Originally, when planning the route, the bicycle may be conceived abstractly as a tool for commuting and compared to other options such as taking the bus, and at this level the mechanistic details of the bicycle and how to ride it may not need to enter into the plan. When the tire breaks, the mechanistic details of how the bicycle works become relevant. Humans navigate such situations naturally, calling to mind the properties of things that are most relevant to the present goal, even as that goal changes on the fly. This again highlights the importance for artificial agents in the human niche to have context- and goal-dependent models of the world and rich ontologies about objects and their properties and affordances.

\subsubsection{Intrinsic value of understanding.}\label{sec:niche-goals-intrinsic-value}

Humans are naturally curious and find it intrinsically rewarding to explore and learn \citep{loewensteinPsychologyCuriosityReview1994}. Being curious and engaging in hypothesis-driven exploration is useful to agents in the human niche for reasons already discussed above. Curiosity is often formalized in the language of reinforcement learning, where agents receive both extrinsic and intrinsic rewards and seek to maximize their sum. If intrinsic reward incentivizes novelty, then in the absence of external rewards, taking actions to maximize reward will result in the agent seeking out novel experiences \citep{schmidhuberPossibilityImplementingCuriosity1991a}. Indeed, there is evidence from neuroscience that extrinsic and intrinsic rewards are processed in similar ways --- or at least by highly overlapping circuits --- in humans and other animals \citep{kaplanSearchNeuralCircuits2007,kiddPsychologyNeuroscienceCuriosity2015,Gottlieb2018,cerveraSystemsNeuroscienceCuriosity2020}. However, novelty means unpredictability and uncertainty, and uncertainty is not necessarily useful. A more nuanced view of curiosity and reward in cognitive science suggests that humans are intrinsically motivated by how quickly their uncertainty is reduced more than by the uncertainty itself (see \cite{poliCuriosityDynamicsOptimal2024} for a recent review).

Uncertainty and learning --- and, therefore, curiosity --- exist at multiple different levels of abstraction \citep{fristonActiveInferenceCuriosity2017}. Uncertainty about the current state of the world motivates a curious agent to make new observations, look beneath rocks, and explore hidden areas. Uncertainty about the future state of the world motivates a curious agent to form plans and contingencies. Finally, uncertainty about the causal dynamics of the world (i.e. uncertainty about the constituents and parameters of a causal model) motivates a curious agent to form and test hypotheses. 

\subsubsection{Social dynamics: demonstration, cooperation, and competition.}\label{sec:niche-goals-social-cooperation}

Observing others' actions and results sometimes is an effective way to learn about the world (Section \ref{sec:niche-environment-others}). However, this goes beyond simply learning about the dynamics of the world by observing the outcomes of others' interventions. By observing others' actions, humans also infer their intent and their state of knowledge about the world \citep{bakerActionUnderstandingInverse2009,ullmanHelpHinderBayesian2009}. Reviewing the ways in which social factors influence how children do causal learning, \citeauthor{gopnikScientificThinkingYoung2012} writes, ``very young children and even infants are sensitive to the intentions of others, particularly their intention to teach, and may draw different conclusions from the evidence that teachers give them than from the evidence they gather themselves.'' The capacity to learn by observing others' actions is amplified by also understanding others' intentions, and those intentions may themselves be rich and compositional \citep{Velez-ginorio2016,davidsonGoalsRewardProducingPrograms2024}.

Social cooperation and competition can make causal learning easier when other agents' intentions are understood \citep{goodmanCauseIntentSocial2009}. But in cases where intentions are not well understood, both cooperation and competition are forms of confounding. In the case of cooperation, the environment can seem causally beneficial if a helper's intent is not accounted for, e.g. in cases where we allow our own children to win games (potentially leading them to wrongly assess their skills). In case of competition, it wrongly seems causally harmful. But in either case, not modeling the confounding intent of others would lead to wrong conclusions about the way the world works. This suggests that some form of ``theory of mind'' is essential for agents in the human niche to learn good (non-confounded) causal models.

\subsubsection{Social dynamics: language.}\label{sec:niche-goals-social-language}

Humans acquire causal knowledge not only through direct experience, but also through linguistic instruction, explanation, and storytelling. Language serves both as a source of causal information and as a cognitive tool for organizing and expressing causal knowledge. A trusted teacher can provide causal knowledge through descriptions that that is just as useful as demonstration or direct experience. Similarly, large corpora of language data contain within them causal descriptions which modern language models are beginning to access and leverage \citep{kicimanCausalReasoningLarge2024,wangCausalBenchComprehensiveBenchmark2024}.

Beyond transmitting information, language also provides structure for causal thought through storytelling. Humans construct narratives that impose coherence on complex events, linking causes and effects across social, physical, and psychological domains. For example, when explaining a car accident, one might link the driver's actions, environmental conditions, and mechanical failures to create a comprehensive narrative. Similarly, in social situations, humans weave causal stories that connect behaviors with motivations, emotions, and social contexts to explain our own and others' actions and intentions. Such narratives integrate multiple levels of causal explanation --- intentions, mechanisms, actions, and outcomes --- into unified accounts \citep{lombrozoStructureFunctionExplanations2006,brunerNarrativeConstructionReality1991}.

Interestingly, forming post hoc (counterfactual) explanations is often easier than generating a priori predictions, perhaps because explanations are grounded in known outcomes and draw on structured causal schemas \citep{wellmanCausalReasoningInformed2007}. This suggests that language-based explanations do not merely describe causality but help to construct it, providing a bridge between abstract causal models and real-world reasoning.

\subsubsection{Social dynamics: credit and blame.}\label{sec:niche-goals-social-credit-blame}

Humans often use their causal knowledge retrospectively, attributing positive outcomes to specific individuals or their actions, or holding them responsible for negative results. This attribution of credit or blame is shaped by an understanding of the individual's intentions and their knowledge at the time they acted. The process involves counterfactual thinking: credit is given where \emph{if actions were different}, the outcomes would have been unfavorable. Blame is attributed where outcomes would be improved if those actions or decisions had not been made \citep{lagnadoCausationLegalMoral2017}. The assignment of credit and blame plays a crucial role in personal learning and in helping groups uphold trust and norms.

\section{A Critical Review of Structural Causal Models}\label{sec:scms}

Structural Causal Models (SCMs) are not the only mathematical formalism for causality, nor are they necessarily the preferred mode of thinking about causality in philosophy, medicine, economics, or the social sciences \citep{cartwrightCausationOneWord2004,psillosCausalPluralism2009,imbensCausalInferenceStatistics2015}. However, they have emerged as the \textit{de facto} standard in machine learning \citep{petersElementsCausalInference2017,kaddourCausalMachineLearning2022}, likely due to their quantitative precision and close links to probabilistic graphical models. In this section, we will review the central assumptions that SCMs make about the world, how they benefit the modeler, and how they sometimes go wrong in the human niche.

\begin{table}[h!]
  \begin{center}
    \label{tab:scm-assumptions}
    \begin{tabular}{P{5cm}|P{4.9cm}|P{5.8cm}}
      \textbf{Assumption / practice} & \textbf{What it lets us do} & \textbf{How it goes wrong} \\ \hline
        \textbf{Tabular data:} all relevant $\lbrace{\rvx_i}\rbrace$ have known semantics and are measured. & Reduce learning to discovery and identification. & Requires additional representation{-\allowbreak}learning in ``raw data'' settings (e.g. processing pixels). \\ \hline
        \textbf{Reichenbach principle:} if $\rvx \not\indep \rvy$, then $\rvx \rightarrow \rvy$, $\rvy \rightarrow \rvx$, or $\rvx \leftarrow \rvz \rightarrow \rvy$. & Propose causal relations from statistical relations. Apply faithfulness conditions. & Not all relations can be parsimoniously described as directed causal relations. \\ \hline
        \textbf{Directed Acyclic Graphs} (DAGs). & Efficient inference. Inductive bias for causal learning. & Awkward to model bidirectional influences and non-causal relations. Less expressive than causal programs. \\ \hline
        \textbf{Discovery and identification:} assume goal of learning is to recover ``true'' model. & Provable guarantees (if data match assumptions). & The ``best'' model depends on modeler's goals and constraints. \\ \hline
        \textbf{Independent exogenous noise:} assume $p(\noise_1, \ldots, \noise_N) = \prod_{i=1}^N p(\noise_i)$ & Equivalent to assuming no unobserved confounders. Counterfactuals$^\star$. & Restricts types of systems that can be modeled and places high demands on observation. \\ \hline
        \textbf{Deterministic functions $\f_i$:} attribute all uncertainty to $\noise$. & More precise model identification. Counterfactuals$^\star$.  & Restricts types of systems that can be modeled and places high demands on fidelity of simulations. \\ \hline
        \textbf{Interventionally Independent Causal Mechanisms (I-ICM):} $\f_i$ is unaffected by intervening on $\f_j$. & Formalize interventions. Do-calculus. Learning through interventions. & Not always feasible to intervene on one mechanism without affecting others. \\ \hline
        \textbf{Statistically Independent Causal Mechanisms (S-ICM):} $\f_i$ is statistically independent of $\f_j$ & Inductive bias for causal learning, e.g. via algorithmic independence. & Ontologically related entities share causal properties. \\ \hline
    \end{tabular}
    \caption{Core assumptions in the SCM framework, what they let us do, and how they sometimes go wrong. See the main text for details. $^\star$ Note: deterministic mechanisms and independent exogenous noise are not strictly necessary for counterfactuals (see Sec. \ref{sec:scm-counterfactuals}).}
    \label{tbl:scm-assumption}
  \end{center}
\end{table}

\begin{figure}
    \centering
    \includegraphics{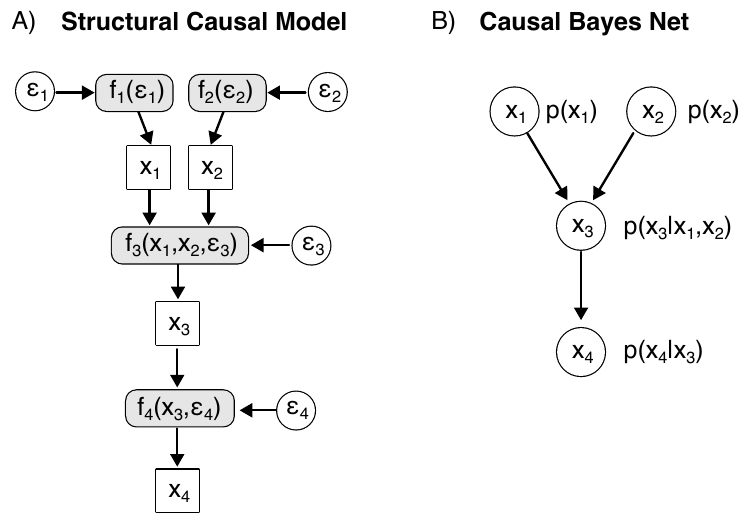}
    \caption{\textbf{A)} A Structural Causal Model (SCM) expresses causal dependencies between variables $\lbrace{\rvx_1,\rvx_2,\rvx_3,\rvx_4}\rbrace$ in terms of deterministic functions $\lbrace{\f_1,\f_2,\f_3,\f_4}\rbrace$ and unknown or variable exogenous noise sources $\lbrace{\noise_1,\noise_2,\noise_3,\noise_4}\rbrace$. \textbf{B)} Causal Bayes Nets (CBNs) are a closely-related type of probabilistic graphical model that expresses causal relations between parents and children in terms of conditional probabilities.}
    \label{fig:scm-cbn}
\end{figure}

\subsection{What is causality and how do Structural Causal Models formalize it?}

This section provides only brief definitions and assumes some familiarity with SCMs and how they are used in ML. For a comprehensive and formal introduction to SCMs, we refer the reader to existing textbooks and tutorials \citep{pearlCausality2009,petersElementsCausalInference2017} and reviews of their application in machine learning \citep{scholkopfCausalityMachineLearning2019,bareinboimCausalReinforcementLearning2020,kaddourCausalMachineLearning2022}.

SCMs generalize probabilistic graphical models, making causal relations and mechanisms explicit \citep{pearlCausality2009,petersElementsCausalInference2017}. SCMs begin by partitioning states of the world into endogenous variables $\lbrace{\rvx_1, \ldots, \rvx_N}\rbrace$ that are part of the system under study, and other exogenous variables $\lbrace{\noise_i, \ldots, \noise_N}\rbrace$ that are outside the system and considered to be ``noise.'' In a SCM, each endogenous variable $\rvx_i$ depends on both its causal parents in the system $\parents_i$ and some exogenous noise $\noise_i$ through a function $\f_i$:
\begin{equation}\label{eqn:scm}
    \rvx_i  = \f_i(\parents_i, \noise_i) \, .
\end{equation}
The function $\f_i$ is referred to as the mechanism relating causes $\parents_i$ to effects $\rvx_i$, and is assumed to be deterministic. This relation is naturally asymmetric ($\parents_i \rightarrow \rvx_i$), and a collection of these relations is described using a directed graph. It is typical to further assume or restrict the graph to be a directed acyclic graph (DAG), although extensions of SCMs to cyclic cases exist \citep{bongersFoundationsStructuralCausal2021}.

Part of the appeal of SCMs is that they express probabilistic relations between variables. Although equation (\ref{eqn:scm}) describes a mechanistic and deterministic relationship between causes $\parents_i$, exogenous noise $\noise_i$, and effects $\rvx_i$, the noise terms $\lbrace{\noise_1,\ldots,\noise_N}\rbrace$ are assumed to be unknown. Marginalization of unknown noise terms results in a probabilistic relationship between causes $\parents_i$ and effects $\rvx_i$, expressed as $p(\rvx_i|\parents_i)$. When a set of variables $\lbrace{\rvx_1, \rvx_2, \ldots, \rvx_N}\rbrace$ is related by structural equations such as (\ref{eqn:scm}), their joint probability distribution can be written according to the disentangled factorization
\begin{equation}\label{eqn:disentangled}
    p(\rvx_1, \rvx_2, \ldots, \rvx_N) = \prod_{i=1}^N p(\rvx_i | \parents_i) \,
\end{equation}
\citep{pearlCausality2009,petersElementsCausalInference2017}. Interpreted as a probabilistic graphical model, Equation (\ref{eqn:disentangled}) can be seen as a causal kind of Bayesian network, or Causal Bayes Net (CBN) (Figure \ref{fig:scm-cbn}). Thus, SCMs generalize probabilistic graphical models and so they can at least do anything that probabilistic models can do, such as inferring from the fact that the ground is wet that it probably rained. SCMs and CBNs add \emph{causal} functionality, above and beyond what can be expressed in classic probabilistic graphical models, by further distinguishing between observation and intervention, as described below.

Another assumption hiding in the above equations is that the exogenous noise terms are independent of each other, or
\begin{equation}\label{eqn:indep-noise}
    p(\noise_1, \ldots, \noise_N) = \prod_{i=1}^N p(\noise_i) \, .
\end{equation}
This assumption is required in order to translate from the structural equations (\ref{eqn:scm}) into the disentangled factorization (\ref{eqn:disentangled}). If, instead, there were additional dependencies between exogenous noise terms $\noise_i \not\indep \noise_j$, then this would in general induce additional dependence between the endogenous variables $\rvx_i$ and $\rvx_j$ that is not accounted for by the mechanisms of the model. Conversely, the assumption that \emph{statistical} independence relations reflect the underlying \emph{causal} mechanisms --- the so-called faithfulness assumption --- is a cornerstone of many causal learning algorithms that seek to learn causal relations from observational data \citep{petersElementsCausalInference2017,zhangThreeFacesFaithfulness2016}. When it comes to modeling the real world, assuming independence among unobserved or exogenous factors of variation (the ``no unobserved confounders'' assumption) as in (\ref{eqn:indep-noise}) should be viewed as a modeling approximation or a convenient fiction.

\subsection{A note on directed asymmetric relations}

\begin{figure}
    \centering
    \includegraphics{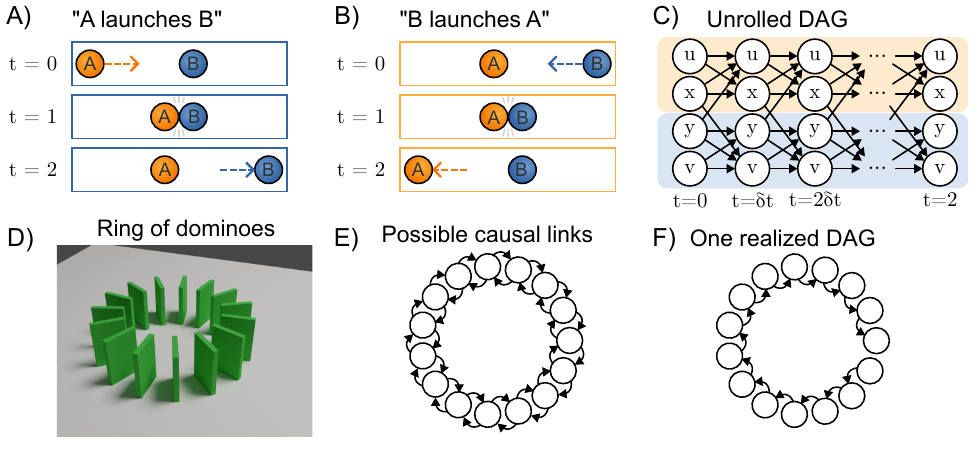}
    \caption{Two simple examples where causal intuitions strain the SCM formalism.
    \textbf{A--C)} Example 1: An elastic billiard ball collision (example inspired by \cite{chengCausalInvarianceEssential2017}).
    \textbf{A)} From the reference frame of ball B, ball A ``launches'' ball B (or, ball B ``stops'' ball A).
    \textbf{B)} The same event viewed from a reference frame initially moving with ball A. Here, ball B ``launches'' ball A (or, ball A ``stops'' ball B).
    \textbf{C)} An accurate but cumbersome way to force directed acyclic graph (DAG) structure onto this symmetric event is to unroll the physical dynamics ($x,u$ = position, velocity of ball A, $y,v$ = position, velocity of ball B).
    \textbf{D--F)} Example 2: A ring of dominoes (example inspired by \cite{cohenGroundedTheoryCausation2022}).
    \textbf{D)} A set of dominoes intuitively supports a series of interventional (``what if I knock over this one?'') or counterfactual (``what if I had instead glued this one down?'') questions.
    \textbf{E)} Each domino is poised to causally influence either of its two neighbors.
    \textbf{F)} Knocking over a particular domino in a particular direction kicks off a chain of events, realizing one particular directed acyclic structure.
    }
    \label{fig:billiards-and-dominoes}
\end{figure}

When it rains the ground gets wet, but soaking the ground with a hose produces no rainclouds. This is a familiar and intuitive causal relation in which one variable (rain) causally influences another variable (wet ground), but not the other way around \citep{pearlCausality2009}. This is the key idea behind using \emph{directed} relations to model causality, as in $\text{rain} \rightarrow \text{wet ground}$ but not $\text{wet ground} \rightarrow \text{rain}$. One puzzle of causality is how symmetric and time-reversible physics on a microscopic scale can give rise to this apparent macroscopic asymmetry of cause and effect. A number of authors have attacked this problem from first principles, explaining how causal asymmetry and the thermodynamic ``arrow of time'' are compatible with symmetric or coupled dynamical systems models of physics \citep{allahverdyanRelatingThermodynamicArrow2008,scholkopfCausalityMachineLearning2019,papineauStatisticalNatureCausation2022}. This line of work --- showing formal conditions where asymmetry is a veridical description of how the macroscopic world really behaves --- is a valid way to approach the apparent asymmetry of causality. However, cases where the world \emph{really is} causally asymmetric are the exception not the rule, so weakening the assumption of true asymmetry would broaden the scope of where we can apply our formal models.

A different approach, more in line with our view that causality is operational, contextual, and subjective, would be to ask a question of human psychology: do humans engage in asymmetric causal reasoning in situations with clear physical symmetry? The answer is clearly yes. In fact, \citeauthor{michottePerceptionCausality2017}'s early experiments on the perception of causality involved situations like elastic collisions between two objects, which is both symmetric and time-reversible (Figure \ref{fig:billiards-and-dominoes}A-C). As argued by \citeauthor{chengCausalInvarianceEssential2017}, even a simple intuitively causal situation like one billiard ball colliding with and  launching another admits different causal interpretations depending on one's frame of reference. And this frame of reference may depend strongly on the goals or vantage point of an observer. For instance, one observer may interpret the scene as ``ball A launches ball B'' if they interpret A as enacting an intervention on B by an agent. Another observer might interpret ball B as causally ``stopping'' or ``impeding'' ball A. Both observers are telling an incomplete but potentially useful story about what happened. Notably, the causal language we use to describe such a situation is always directed and asymmetric, but the direction of cause and effect can flip from one telling to another.

A related example worth highlighting comes from \citeauthor{cohenGroundedTheoryCausation2022}, illustrated in Figure \ref{fig:billiards-and-dominoes}D-F. This is another example drawing on intuitive (Newtonian) physics to tell a simple causal story about interacting object. Imagine a set of dominoes arranged in a way that any one could knock over either of its neighbors, and the whole set forms a cycle. Intervening on this system by picking any one domino and knocking it over in a particular direction results in a chain reaction where all dominoes eventually end up in a fallen-over state, each causally influenced by the one before it in the chain and causally influencing one after. Once this chain reaction is started, it has a nice directed and acyclic structure (Figure \ref{fig:billiards-and-dominoes}F), but critically \emph{which DAG is realized depends on the intervention itself} \citep{cohenGroundedTheoryCausation2022}.

Even simple physical examples like these, which fall well short of the complexity of the real world, frustrate any attempt to capture the intuitive causal dynamics at play with a single SCM. The fact that humans find these examples intuitive further illustrates the importance of modeling the dynamics of the real world with context-, goal-, and even action-dependent causal models.

\subsection{Interventions and modularity}

The ability to model interventions is a crucial aspect of any formal theory of causality. Interventional questions are of the form ``what would happen if...?'', for instance ``what would happen if I turned on the hose?''. In SCMs and CBNs, a key assumption is that the system being modeled is modular, i.e. that interventions can occur locally in the model, affecting a single variable or mechanism or a sparse subset of them. Using the $\sim$ symbol to denote the change due to intervening, an intervention on the variable $\rvx_i$ or mechanism $\f_i$ is expressed formally as
\begin{equation}\label{eqn:interventions}
    p_{\dooperator(\f_i \rightarrow \tilde\f_i)}(\rvx_1, \rvx_2, \ldots, \rvx_N) = \tilde{p}(\vx_i|\tilde\parents_i) \prod_{j\neq i} p(\rvx_j | \parents_j) \, ,
\end{equation}
where $\tilde{p}(\vx_i|\tilde\parents_i)$ is the result of marginalizing over $\noise_i$ with the updated mechanism $\tilde{f}_i$ that has been intervened on.\footnote{We write $\tilde\parents_i$ because in general, interventions may modify the graph in a way that adds or removes parents for $\rvx_i$. An atomic intervention is one where $\rvx_i$ is set to a particular value $\hat\rvx_i$, and can be thought of as a special case of $\tilde\f_i$ where $\tilde\f_i(\parents_i,\noise_i)=\hat\rvx_i$ and $\tilde{p}(\rvx_i|\parents_i) = \tilde{p}(\rvx_i) = \delta(\rvx_i - \hat\rvx_i)$.} 
Equation (\ref{eqn:interventions}) formalizes the idea of local or modular interventions in the sense that the terms in the product over $j \neq i$ are unchanged. Turning a hose on or off not only affects the state of the ground, but importantly it also \emph{does not} have any effect on the way in which rain affects the ground. Local interventions then propagate through the graph, affecting descendants of $\rvx_i$ but none of its non-descendants. Modularity as in equation (\ref{eqn:interventions}) is foundational to the study of causality. For instance, without modularity, there is no do-calculus \citep{pearlCausality2009,petersElementsCausalInference2017}. 

\subsection{Counterfactuals and hidden variables}\label{sec:scm-counterfactuals}

Whereas interventional questions are of the form ``what would happen if...?'', counterfactual questions are of the form ``all else being equal, what would have happened if...?''. Counterfactual questions are essentially interventional questions about how things could have gone differently in the past. The key difference is that counterfactuals can draw on additional information that becomes available after observing a specific outcome. The theory of SCMs has revealed intriguing formal distinctions between interventions and counterfactuals \citep{bareinboimPearlHierarchyFoundations2020a}, though in many ways the practical implications of this distinction are still an open question.

Where interventional statements often apply to a broad population, counterfactual statements can be applied to individual events or cases. For example, consider a situation where a patient is given a new drug and experiences a positive health outcome. A counterfactual question would be ``what would have happened if the patient had not received the drug?'' Crucially, observing the positive health outcome and running additional tests after the fact provides additional information about the patient's condition that was not available at the time that the drug was first administered; for example, we might become more confident in the patient's original diagnosis after seeing how they responded to the drug. In the SCM formalism, a counterfactual statement such as this is formed in three steps \citep{pearlCausality2009,pearlStructuralCounterfactualsBrief2013}:
\begin{itemize}
    \item[(i)] some initial outcome $\lbrace{\hat\rvx_1, \hat\rvx_2, \ldots, \hat\rvx_N}\rbrace$ is observed;
    \item[(ii)] values of unobserved or exogenous terms are inferred, that is, the posterior distribution over the noise terms $p(\noise_1,\ldots,\noise_N|\hat\rvx_1,\ldots,\hat\rvx_N)$ is computed; and
    \item[(iii)] an intervention is quantified just like in equation (\ref{eqn:interventions}) but in a modified model where the independent prior over exogenous noise terms $p(\noise)$ is replaced by inferred values from the posterior $p(\noise_1,\ldots,\noise_N|\hat\rvx_1,\ldots,\hat\rvx_N)$.
\end{itemize}
Intuitively, substituting the posterior over the exogenous noise terms for the prior formalizes the ``all else being equal'' part of the counterfactual query. The posterior distribution over exogenous terms reflects information that is specific to that outcome or individual, while the prior captures the population. This three-step recipe provides a precise quantitative method for quantifying counterfactual statements and has featured prominently in models of human thought and explanation \citep{pearlStructuralCounterfactualsBrief2013,slomanCausalityThought2015}.

A common claim motivating the use of SCMs over CBNs is that while CBNs can model interventions, they cannot model counterfactuals; this effectively separates the second and third levels of the causal hierarchy \citep{bareinboimPearlHierarchyFoundations2020a}. As the story goes, CBNs lack explicit exogenous noise terms to be inferred and transferred to hypothetical alternative scenarios. While this is true in cases where all endogenous variables are observed, it is misleading. The key property of SCMs that allows them to form counterfactuals is that the noise terms are \emph{unobserved}, not that they are independent. Any causal model, including those in the CBN family, that includes unobserved terms has the ability to form counterfactuals by following the three-step procedure outlined above: substituting the posterior over unobserved terms for the prior and then running the usual machinery of interventions. There is no strict requirement that conditional probabilities are constructed by deterministic mechanisms, nor that unknown quantities are \textit{a priori} independent of each other and ``exogenous,'' although these are certainly convenient assumptions to make for other reasons.

Humans engage in counterfactual reasoning all the time, and earlier we discussed a few ways in which, in the human niche, this is an adaptive strategy for dealing with misspecified models (Sec \ref{sec:niche-constraints-models-wrong}) and the social dynamics of credit and blame (Sec. \ref{sec:niche-goals-social-credit-blame}). Countefactuals can be implemented in a variety of different formalisms, such as causal programs \citep{tavaresLanguageCounterfactualGenerative2021} or CBNs with unobserved variables.

\subsection{Learning: where to SCMs come from?}

Before one can \emph{use} a causal model --- i.e. query it about interventions or counterfactuals --- one needs to first write down the model, but in many cases we are interested in, the model is not known ahead of time and must be learned. Within the SCM framework, the problem of learning a causal model from data is typically broken down into the sub-problems of causal discovery, identifying which arrows in the graph exist, and causal identification, finding the specific functional form of each $\f_i$ and $p(\noise_i)$. A model is said to be identifiable from some data-generating process if, in the limit of more and more data, learning converges to the true model or an equivalent model. With little prior knowledge, causal discovery is challenging because the number of possible directed graphs for $N$ variables is super-exponential in $N$ \citep{spirtesCausationPredictionSearch2000,petersElementsCausalInference2017}. Further, discovering and identifying a causal model from merely observational data is not possible in general \citep{pearlCausality2009,petersElementsCausalInference2017,xiaCausalNeuralConnectionExpressiveness2021}. These are interesting computational challenges, and the problem of discovering and identifying causal models from data has captured the interest of many machine learning researchers.

Causal statements like ``smoking increases the risk of lung cancer'' are coarse-grained statistical statements; ``smoking'' and ``lung cancer'' each describe some large set of possible world states. This raises the question: for any given problem, where do the variables themselves come from, and how many of them are there (what is $N$)? In tabular settings, an expert has pre-defined a set of $N$ variables of potential interest: a physician might prescribe a particular drug and monitor a particular set of biomarkers, an economist might monitor quantities like GDP, taxation, and welfare, and a social scientist might survey a population with a carefully designed questionnaire. In all of these tabular settings, there is a relatively clear separation between the endogenous variables of interest and the exogenous ``noise'' outside the scope of a particular model. Domain experts may have knowledge that licenses an assumption that there are no unobserved confounders. In such settings, learning a causal model involves only discovery and identification. As such, SCMs have been highly impactful tools in tabular settings such as those common in medicine, economics, and social sciences. However, outside the tabular setting, such as learning a causal model of the world from video or other raw sensory data, a partitioning of the world into $N$ endogenous variables such that exogenous terms are statistically independent is not given ahead of time and must itself be learned; this is the causal representation learning problem \citep{scholkopfCausalRepresentationLearning2021,beckersApproximateCausalAbstraction2019,holtgenEncodingCausalMacrovariables2021}.

Even within the SCM formalism, learning a causal model from data is computationally hard and sometimes intractable. As a general rule, one can make progress by making assumptions about the structure of the model (e.g. sparsity), by making assumptions about the data-generating process (e.g. which variables are intervened on), or both. Much of the literature on causality, especially in conjunction with machine learning, is dedicated to the noble pursuit of identifiability: finding a small number of reasonable assumptions about model structure and the data-generating process such that, in domains that conform to those assumptions, the true model is guaranteed to be recovered given enough data. However, as we argued above, discovering the ``true'' model and learning a ``useful'' model are related but distinct goals.

In the following subsections, we will discuss assumptions having to do with the model structure and, in turn, the structure of the data-generating process.

\subsection{Assumptions on model structure}\label{sec:scm_model_structure}

A common approach to simplifying the causal learning problem is to make assumptions about the structure or functional form of the SCM that restrict the class of possible SCMs and thus improve identifiability \citep{petersElementsCausalInference2017}. We have already seen a few assumptions about the structure of the model: equations (\ref{eqn:scm}) and (\ref{eqn:indep-noise}) assume deterministic mechanisms and independent exogenous noise for each variable, and the resulting graph is static and is typically constrained to be acyclic. Other common assumptions in this vein include sparse or low-degree connections \citep{claassenLearningSparseCausal2013}, additive noise \citep{petersCausalDiscoveryContinuous2014}, or non-Gaussian noise \citep{hyvarinenNonlinearIndependentComponent2023}. While increasing identifiability leads to better guarantees on the outcome of the learning process, it is important to remember that learning is in service of deployment, and recovering the ``true'' model is not always necessary to get good-enough answers to interventional and counterfactual queries (and it may not be possible to define the ``true'' model in real world settings). A more general approach is to seek a collection of plausible models, i.e. to model uncertainty over the model itself and apply Bayesian inference to the discovery and identification problems \citep{deleuBayesianStructureLearning2022}. Interventional and counterfactual queries can then be averaged over all plausible models, providing quantitative answers and error bounds even in the absence of guarantees on identifiability.

Models are always a simplification of the system they describe, and so one should always be careful to distinguish true properties of the world from useful properties of the model. The assumption that exogenous noise is \textit{a priori} independent (equation (\ref{eqn:indep-noise})) is usually justified by an appeal to Reichenbach's principle, which states that two variables $\rvx$ and $\rvy$ are statistically dependent only if one of three cases is true: either $\rvx \rightarrow \rvy$, $\rvy \rightarrow \rvx$, or $\exists \rvz: \rvx \leftarrow \rvz \rightarrow \rvy$ \citep{petersElementsCausalInference2017}. While this principle expresses a fundamental truth about the universe --- correlations exist when some underlying physical processes connect $\rvx$ to $\rvy$ --- this does not imply that all correlations are usefully \emph{modeled} as such (Sec. \ref{sec:niche-environment-complex}). Further, it is not always feasible to observe all relevant variables. If $\rvz$ is a common cause of $\rvx$ and $\rvy$, then $\rvz$ being unobserved induces a dependence between the noise term for $\rvx$ and the noise term for $\rvy$ (Sec. \ref{sec:niche-environment-confounded}). The exogenous noise terms in SCMs are a particular kind of latent or unobserved variable, and the assumption in equation (\ref{eqn:indep-noise}) that they are independent of each other (the no unobserved confounders assumption) is not just a restriction on the kinds of system we can effectively model, but also places strong demands on the observations we make of those systems (Sec. \ref{sec:niche-constraints-observations}).

Another fundamental structural assumption of SCMs is known as the Independent Causal Mechanisms (ICM) principle. The ICM principle comes in two flavors that should be treated as separate principles; we will refer to them as Interventionally Independent Causal Mechanisms (I-ICM) and Statistically Independent Causal Mechanisms (S-ICM). The I-ICM principle states that an intervention on mechanism $\f_i$ has no effect on other mechanisms $\f_j$, which captures the idea that interventions are \emph{local} and causal systems are \emph{modular}. The I-ICM principle is the reason that equation (\ref{eqn:interventions}) can be written with only a single term changed and the other $N-1$ factors unchanged. Thus, the I-ICM assumption is foundational for causal thinking and formalizing what it means to ``intervene'' on a system \citep{woodwardMakingThingsHappen2005} and its implications for SCMs are well understood. However, critics have pointed out that many real-world systems admit intuitive and useful (to humans) decomposition into individual mechanisms or components that are far from modular; there may be no physically possible intervention that targets a single mechanism \citep{cartwrightCausationOneWord2004}.

The S-ICM principle, on the other hand, is about statistical independence of mechanisms. Basically, it states that the mechanisms $\f_i$ and $\f_j$ are drawn \textit{a priori} independently from a pool of possible mechanisms. This is sometimes expressed in terms of the algorithmic independence of the functions $\f_i$ and $\f_j$ themselves \citep{janzingCausalInferenceUsing2010,lemeireReplacingCausalFaithfulness2013,petersElementsCausalInference2017}. The S-ICM principle implies that data that provide information about one mechanism $\f_i$ do not provide additional information about other mechanisms $\f_j$. Like the Reichenbach principle, we argue that the S-ICM principle is not always \emph{useful}, regardless of whether it is true in a metaphysical sense. Imagine, for instance, walking into an unfamiliar and dimly lit room and looking for an intervention that would turn on the lights. There is an unfamiliar push button rather than a familiar toggle switch on the wall. Humans readily hypothesize that objects of similar types ($\lbrace{\texttt{push\_button,toggle\_switch}}\rbrace$) may be involved in similar mechanisms ($\lbrace{\texttt{toggles\_lights}}\rbrace$) \citep{griffithsTheorybasedCausalInduction2009,zhaoHowPeopleGeneralize2022}. In other words, causal properties are shared among ontologically similar objects, so learning about one mechanism gives information about related mechanisms. Expressing this in the language of SCMs has been shown to speed up learning \citep{brouillardTypingAssumptionsImprove2022} but requires significantly modifying or abandoning the S-ICM principle.

\subsection{Assumptions on the data-generating process}\label{sec:data_assumptions}

Observational data refers to information that is passively collected without intervention or manipulation. While causal identification from observational data is not possible in general, it is possible in cases where the structure of the graph is known and satisfies certain criteria given by the do-calculus \citep{pearlCausality2009}. On the other hand, learning from direct interventions simplifies causal learning by essentially turning it into a supervised learning problem: predict outcomes from interventions where both are observed. In many settings, interventions require significant involvement and control, which may not always be feasible or ethical. In these settings, it is necessary to address the challenging task of learning causal relations from observational or quasi-experimental \citep{liuQuantifyingCausalityData2021} data. For example, the do-calculus provides methods to quantify the causal impact of smoking on lung cancer in the absence of (infeasible and unethical) randomized control trials \citep{pearlCausality2009}. Conversely, the utility of the do-calculus (and other methods for learning from observational or quasi-experimental data) is reduced in settings where interventions are cheap. Humans, especially young children, learn by doing \citep{schulzLearningDoingIntervention2007}.

Independent and identically distributed (iid) data has the property that each observation is conditionally independent of the other observations given the underlying model or process, and there is just one underlying process generating all of the data. A weaker form of independence is exchangeability, which requires that the particular order of observations is irrelevant, and which has been generalized to the case of causal learning \citep{guoCausalFinettiIdentification2022}. Independence (or exchangeability) is mathematically convenient, but some kinds of non-independence add valuable learning signals. For example, if data come in (non-independent) pairs where the second item is generated by some unknown intervention on the first, identifying the true latent factors of variation becomes possible \citep{shuWeaklySupervisedDisentanglement2020,brehmerWeaklySupervisedCausal2022,ahujaWeaklySupervisedRepresentation2022}. This kind of weakly supervised learning breaks the iid assumption and has been shown to be a powerful tool for causal learning.

\section{Discussion}\label{sec:discussion}

Human cognition is adapted to the world we live in and humans are adept at making sense of the world in causal terms. Although work at the intersection of machine learning and causality has recently flourished, there remain important qualitative gaps between the kinds of causal models developed in ML and what we know about human causal cognition. Closing this gap promises to provide more capable and interpretable ML systems as well as a deeper understanding of human cognition.

We took a critical look at common assumptions in the SCM framework for causality through the lens of modeling ``real world'' causal interactions of the kind we as humans encounter on a daily basis. These critiques do not always apply, and there is no doubt that the SCM formalism is useful and impactful in many important domains such as medicine and economics. The proliferation of SCMs has also increased causal literacy throughout computer science, leading to advances in areas such as fairness, recommender systems, reinforcement learning, and robotics. In short, SCMs are incredibly useful in situations where the SCM assumptions at least approximately hold. 

To take a fresh look at causality and suggest avenues for future work in ML, we examined human causal cognition through an ecological lens. By motivating aspects of human causal cognition by properties of the ``human niche,'' we hope to motivate research and implementation of more causally savvy artificial agents that will be well-adapted to their operating niche. Agents' capacities should be motivated by properties of their environment, constraints on their form and processing power, and their (or their designers') goals. The inverses of these arguments also apply: Agents designed to operate in well-controlled settings, such as factories, will have little use for curiosity, hypothesis-driven exploration, or causal induction, all of which were motivated in part as human adaptations to being generalist agents in an open-ended world.

Multiple themes emerged where some properties of human causal cognition could be seen as adaptations to multiple properties of the human niche. Some of these repeated themes are as follows.
\begin{itemize}
    \item Coarse-graining and abstraction
    \item Sparsity and other kinds of simplifying structure
    \item Zero-shot causal induction from causal theories and ontology
    \item Curiosity and hypothesis-driven exploration
    \item Uncertainty and amortized inference
    \item Epistemic humility and the value of experimentation
\end{itemize}
Further work on these recurring themes will be especially useful to ML in the future.

Although it is important and timely to address this gap between human causal cognition and the way we approach causality in ML, this is far from the only gap between natural and artificial intelligence. In fact, this is another theme that emerged throughout our discussion of the human niche: Human understanding of the world encompasses diverse kinds of modalities and relations, and causal relations are just one part of the story. Taking a step back from causality specifically, we hope that the kind of ecological approach demonstrated here will continue to foster mutually beneficial collaborations between ML and cognitive science.

\section{Bibliography}
\bibliographystyle{myapsr}
\bibliography{references.bib}

\end{document}